\documentclass[a4paper, 10pt, conference]{ieeeconf}      
\usepackage{FG2025}
\usepackage{graphicx}
\usepackage{amsmath}
\usepackage[colorinlistoftodos]{todonotes}
\usepackage{booktabs}
\usepackage{multirow}
\usepackage{acronym}
\usepackage{hyperref}

\newcommand{\defineacronyms}{
  \acrodef{FPR}[FPR]{False Positive Rate}
  \acrodef{TPR}[TPR]{True Positive Rate}
  \acrodef{Flux}[Flux]{FLUX.1-dev}
  \acrodef{SD35}[SD 3.5]{Stable diffusion 3.5 Large}
  \acrodef{EpicRealism}[Epic Realism]{SDXL Epic Realism}
}

\graphicspath{ {./images/} }
\FGfinalcopy 

\IEEEoverridecommandlockouts                              
\def\FGPaperID{****} 

\title{\LARGE \bf
Can Text-to-Image Generative Models Accurately Depict Age? A Comparative Study on Synthetic Portrait Generation and Age Estimation
}

\author{\parbox{16cm}{\centering
    {\large Alexey A. Novikov$^1$, Miroslav Vranka$^2$, François David $^1$, Artem Voronin$^1$}\\
    {\normalsize
    $^1$ Montreal, Canada, $^2$ Presov, Slovakia}}
}

\begin{document}

\defineacronyms

\ifFGfinal
\thispagestyle{empty}
\pagestyle{empty}
\else
\pagestyle{plain}
\fi
\maketitle

\begin{abstract}
Text-to-image generative models have shown remarkable progress in producing diverse and photorealistic outputs. In this paper, we present a comprehensive analysis of their effectiveness in creating synthetic portraits that accurately represent various demographic attributes, with a special focus on age, nationality, and gender. Our evaluation employs prompts specifying detailed profiles (e.g., “Photorealistic selfie photo of a 32-year-old Canadian male”), covering a broad spectrum of 212 nationalities, 30 distinct ages from 10 to 78, and balanced gender representation.
We compare the generated images against ground truth age estimates from two established age-estimation models to assess how faithfully age is depicted. Our findings reveal that although text-to-image models can consistently generate faces reflecting different identities, the accuracy with which they capture specific ages—and do so across diverse demographic backgrounds—remains highly variable. These results suggest that current synthetic data may be insufficiently reliable for high-stakes age-related tasks requiring robust precision, unless practitioners are prepared to invest in significant filtering and curation. Nevertheless, they may still be useful in less sensitive or exploratory applications, where absolute age precision is not critical.
\end{abstract}


\section{INTRODUCTION}

Generative AI text-to-image frameworks (e.g., DALL·E, Midjourney, Stable Diffusion) have recently gained significant traction for their ability to synthesize highly realistic images based on textual prompts. Despite their remarkable capacity to capture distinct facial features or styles, questions remain regarding their utility in specialized biometric tasks, most notably age estimation and age-specific portrait generation. The ability to accurately represent a particular individual at a specified age may yield substantial benefits for data augmentation, privacy-preserving research, and controlled scenario testing.

However, the degree to which these generative models can reflect nuanced facial aging patterns or reliably render subtle age indicators (e.g., skin texture, wrinkles, or the shape of facial landmarks) has not been sufficiently explored. Moreover, relying on synthetic images for tasks such as training age-estimation models introduces critical questions related to domain gaps, data quality, and bias. This paper aims to demonstrate empirically that while generative approaches excel at maintaining identity consistency, their performance in accurately depicting specific ages lags behind real-world image data.

In the following sections, we describe previous work on generative models for facial analysis~\ref{sec:related_work}, present our approach to evaluating and benchmarking age-specific image generation~\ref{sec:methodology}, and discuss our experimental results alongside current limitations and future considerations~\ref{sec:results}.

\section{RELATED WORK}
\label{sec:related_work}

Face aging research focuses on transforming a person’s face to predict how they might look at different ages. Most approaches rely on generative models, particularly Generative Adversarial Networks (GANs)~\cite{goodfellow2014}. While these models produce compelling results, they often struggle to preserve identity details throughout the aging process. To address this, Wang et al.~\cite{tang2018} proposed IPCGANs, which incorporate identity-preserving constraints by ensuring output images retain essential CNN-extracted features of the input face.

A persistent challenge in face aging is the reliance on age-labeled datasets, which frequently exhibit skewed age distributions that undermine both accuracy and realism. Chen et al.~\cite{chen2023} introduced FADING to mitigate this issue using diffusion-based models and large language–image pretraining. Although FADING reduces reliance on biased datasets, it can still inherit biases from the underlying diffusion model. Despite this limitation, FADING achieves more realistic facial transformations and better identity preservation.

For 3D-aware face aging, Wahid et al.~\cite{wahid2024} proposed a two-step pipeline. First, they generate multi-view synthetic aging data using StyleGAN2~\cite{Karras2019}, a CLIP-based~\cite{radford2021} age guidance mechanism, and a 3D-aware generator for consistent aging across various angles. 

Although synthetic face-aging techniques have shown promise in enhancing age-invariant facial recognition~\cite{yao2024}, there is comparatively little work exploring whether generative approaches genuinely benefit age estimation or regression within biometric contexts. Existing studies largely concentrate on aging transformations rather than evaluating how such synthetic data might improve the performance of age estimation models. In this paper, we aim to address these gaps by systematically examining how generative text-to-image models represent diverse demographic attributes—including age, nationality, and gender—and assessing the reliability of these synthetic portraits for training and benchmarking in age estimation tasks. Our findings, supported by extensive evaluations across 212 nationalities, 30 different ages, and balanced gender representation, highlight both the potential and the limitations of current generative techniques in accurately depicting age across varied demographics.

\section{METHODOLOGY}
\label{sec:methodology}

Our goal is to evaluate whether existing text-to-image (T2I) generative AI models can reliably render human faces at specific, user-defined ages. Below, we outline the key components of our methodology:

\subsection{Dataset and Prompt Collection}

\subsubsection{Prompt Design}

We designed a comprehensive set of prompts to systematically evaluate the performance
of generative models across a spectrum of demographic and situational contexts.
Each prompt included exactly one nationality, age, and gender, formatted as follows:

\begin{quote}
\textbf{\emph{``Photorealistic selfie photo of a [age]-year-old [nationality] [gender] person, centered, high-resolution.''}}
\end{quote}


We included 212 nationalities to ensure global coverage, with each nationality represented by 60 to 120 prompts, thereby achieving a balanced representation across different cultural contexts. Age diversity was captured by including 30 distinct ages from 10 to 78 years old, with each age having 432 associated prompts. Gender was evenly split, with 6,480 prompts for each of "male" and "female." This methodology facilitated a detailed examination of how these models handle demographic diversity, providing insights into issues of fairness, bias, and the accuracy of representation in generated images.

\begin{figure*}[t!]
    \centering
    \begin{minipage}{0.16\textwidth}
        \centering
        \includegraphics[width=\textwidth]{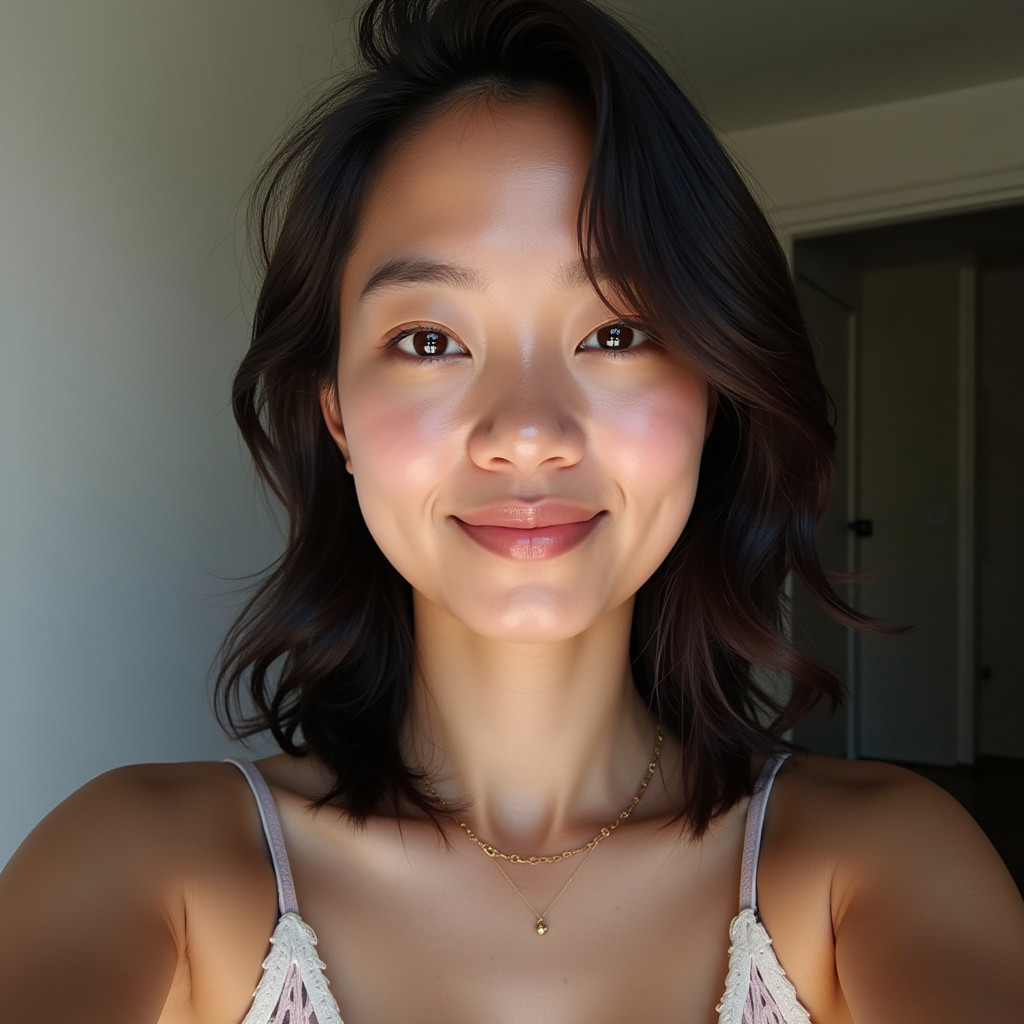}
        (a)
    \end{minipage}\hfill
    \begin{minipage}{0.16\textwidth}
        \centering
        \includegraphics[width=\textwidth]{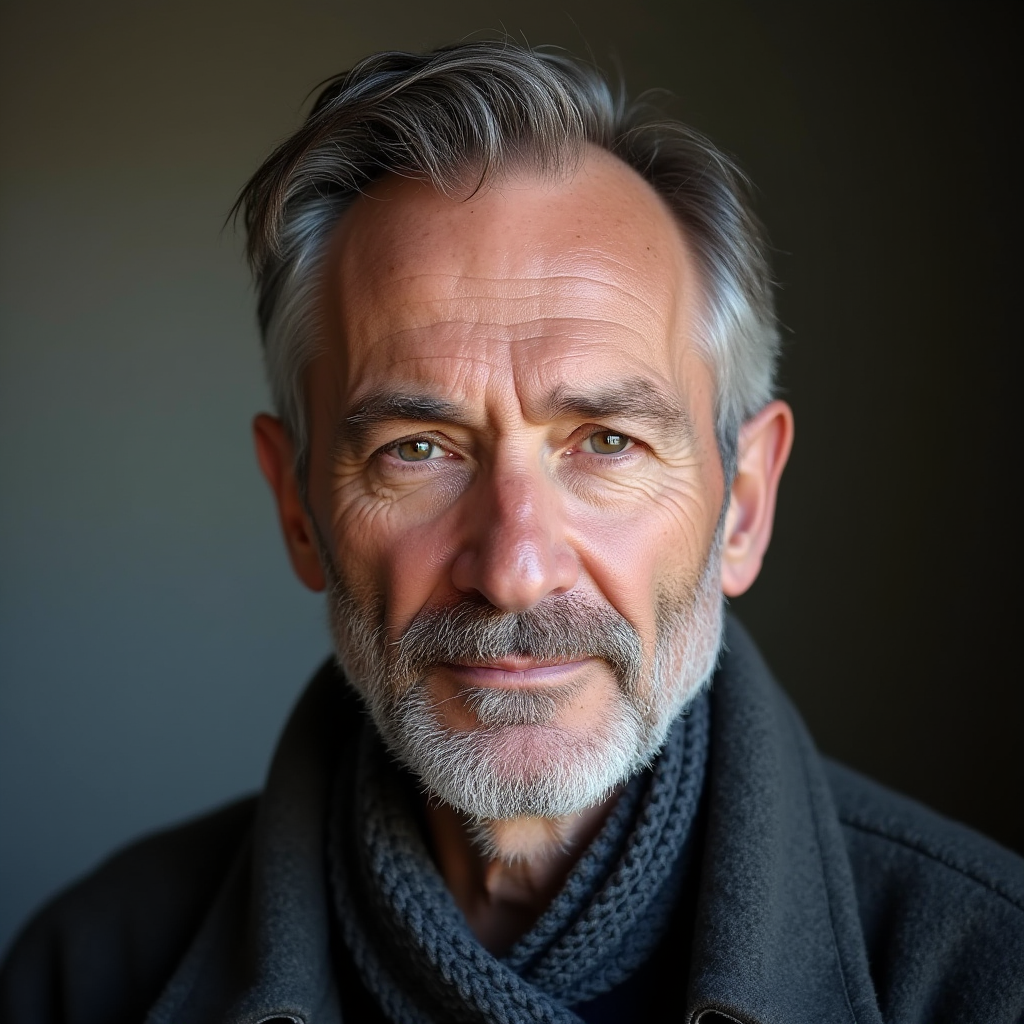}
        (b)
    \end{minipage}\hfill
    \begin{minipage}{0.16\textwidth}
        \centering
        \includegraphics[width=\textwidth]{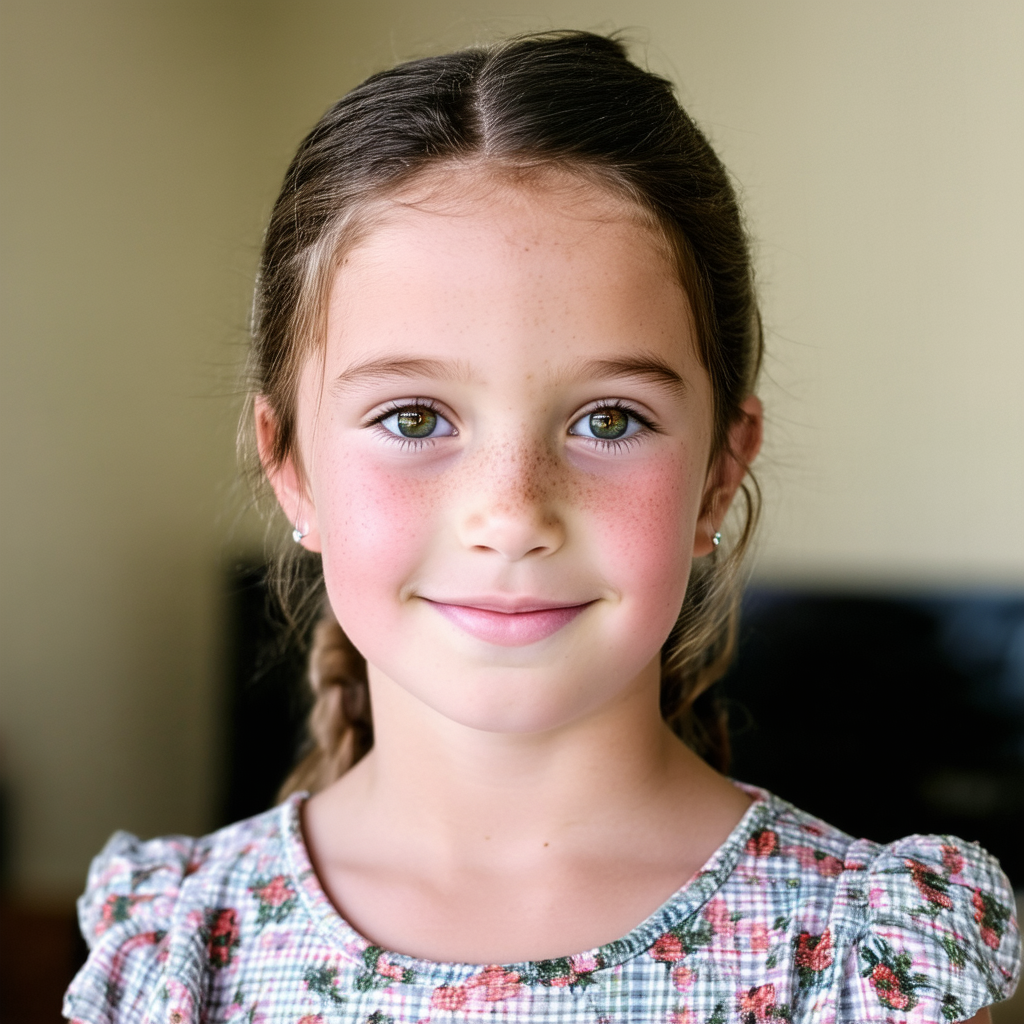}
        (c)
    \end{minipage}\hfill
    \begin{minipage}{0.16\textwidth}
        \centering
        \includegraphics[width=\textwidth]{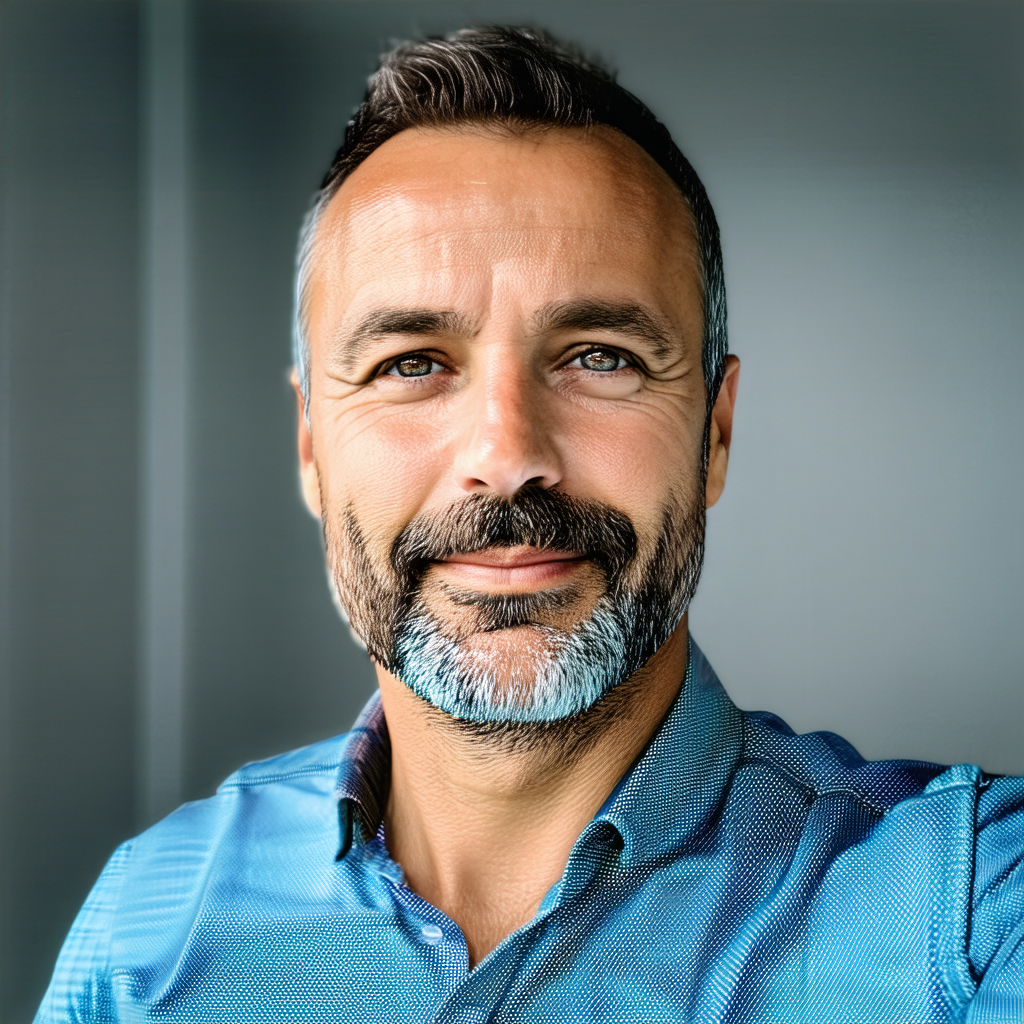}
        (d)
    \end{minipage}\hfill
    \begin{minipage}{0.16\textwidth}
        \centering
        \includegraphics[width=\textwidth]{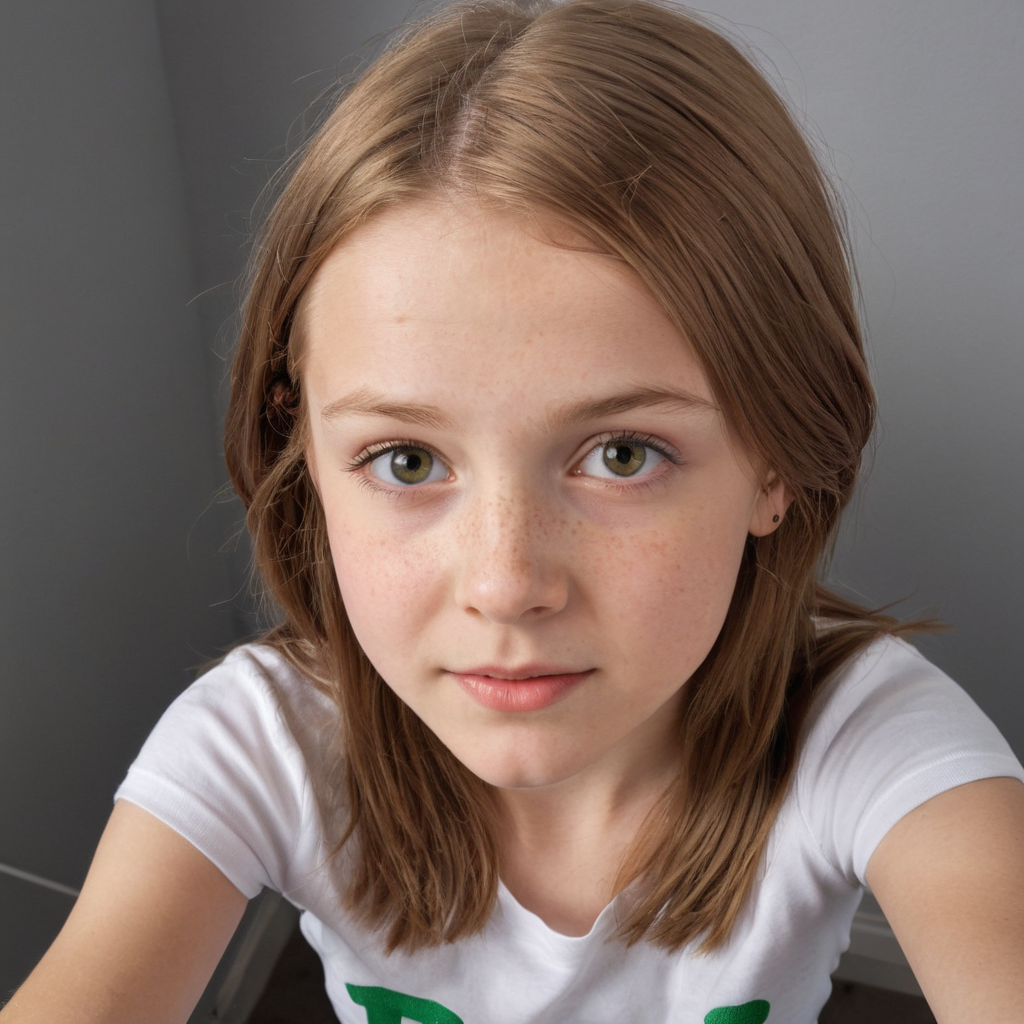}
        (e)
    \end{minipage}\hfill
    \begin{minipage}{0.16\textwidth}
        \centering
        \includegraphics[width=\textwidth]{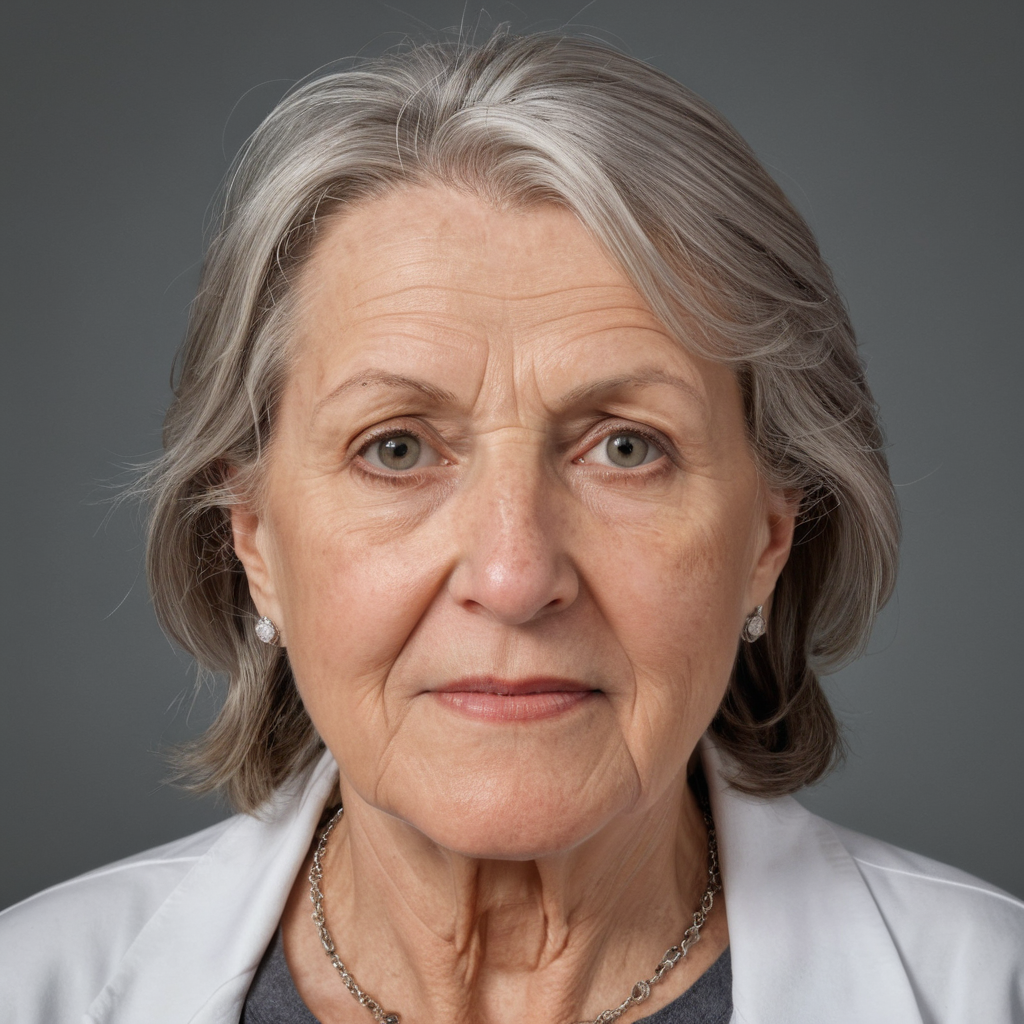}
        (f)
    \end{minipage}\hfill
    \caption{Examples of generated images using different methods: Flux (a, b), Stable Diffusion 3.5 Large (c, d), and SDXL Epic Realism (e, f).}
    \label{fig:examples}
\end{figure*}

\subsubsection{Generative Models}
We employed three text-to-image models \ac{Flux}~\cite{flux_dev}, \ac{SD35}~\cite{sd_35_large}, and \ac{EpicRealism}~\cite{sdxl_epic_realism}—to generate images for each of the 12,960 prompts. All three models are widely recognized for their advanced visual fidelity and nuanced control over image generation, providing a robust baseline for quality. By using these models, we could explore how different architectures and sampling techniques influence output quality and stylistic expression across a broad range of parameter settings. In addition to positive prompts, we employed manually crafted negative prompts to suppress common generative issues, such as distorted faces or extra limbs, thereby improving overall realism. 
In all three models, we kept the output dimensions at 1,024 pixels in both width and height.

Below, we summarize the core configurations for each model:
\paragraph{\acs{Flux}}
We employed this model to generate high-resolution outputs using a moderate number of inference steps and a balanced guidance factor. A basic scheduler in conjunction with an Euler sampling method provided stable results, while shifting parameters refined image fidelity. Random noise seeds and a randomization mechanism ensured each run had controlled variability. The complete configuration is outlined in Table~\ref{tab:fluxdevparams}.

\begin{table}[h]
\centering
\begin{tabular}{ll}
\toprule
\textbf{Parameter} & \textbf{Value} \\
\midrule
\texttt{unet\_name} & \texttt{flux-l.dev.safetensors} \\
\texttt{steps} & 20 \\
\texttt{guidance} & 3.5 \\
\texttt{scheduler} & \texttt{BasicScheduler} \\
\texttt{denoise} & 1.00 \\
\texttt{sampler\_name} & \texttt{euler} \\
\texttt{max\_shift} & 1.15 \\
\texttt{base\_shift} & 0.50 \\
\texttt{control\_after\_generate} & \texttt{randomize} \\
\bottomrule
\end{tabular}
\caption{Key parameters for the \acs{Flux} model.}
\label{tab:fluxdevparams}
\end{table}

\paragraph{\ac{SD35}}
The model was configured to balance generation speed and output detail using a moderate number of inference steps and a moderate classifier-free guidance factor. An Euler sampler and a normal scheduling scheme helped maintain consistency, while a carefully chosen negative prompt mitigated artifacts like blurred or distorted features. All relevant parameters appear in Table~\ref{tab:sd35params}.

\begin{table}[h]
\centering
\begin{tabular}{ll}
\toprule
\textbf{Parameter} & \textbf{Value} \\
\midrule
\texttt{steps} & 15 \\
\texttt{cfg} & 4 \\
\texttt{sampler\_name} & \texttt{euler} \\
\texttt{scheduler} & \texttt{normal} \\
\texttt{negative\_prompt} & \emph{``blurry, overexposed, low quality,}\\
 & \emph{cartoonish, distorted face, extra limbs''} \\
\bottomrule
\end{tabular}
\caption{Key parameters for the \texttt{sd\_35\_large} model.}
\label{tab:sd35params}
\end{table}

\paragraph{\acs{EpicRealism}}
Finally, this model was chosen for its refined realism and higher fidelity, guided by a moderate classifier-free guidance setting. It employed a \texttt{dpmpp\_3m\_sde} sampler and an exponential schedule, along with a detailed negative prompt to avoid residual distortions or “uncanny” appearances. Table~\ref{tab:sdxlparams} summarizes the full parameter configuration.

\begin{table}[h]
\centering
\begin{tabular}{ll}
\toprule
\textbf{Parameter} & \textbf{Value} \\
\midrule
\texttt{steps} & 30 \\
\texttt{cfg} & 3.5 \\
\texttt{sampler\_name} & \texttt{dpmpp\_3m\_sde} \\
\texttt{scheduler} & \texttt{exponential} \\
\texttt{negative\_prompt} & \emph{``deformed iris, deformed pupils,}\\
 & \emph{cgi, cartoon, mutated hands''} \\
\bottomrule
\end{tabular}
\caption{Key parameters for the \texttt{sdxl\_epic\_realism} model.}
\label{tab:sdxlparams}
\end{table}


In the Fig.~\ref{fig:examples} we show few examples generated by the models. In the future, once additional refinements are made to remove
remaining noise or possible artifacts, we aim to release the resulting dataset for broader use. 

\subsection{Age Estimation}

We utilized the publicly available age estimation model MiVOLO-D1~\cite{Kuprashevich2023},
recognized as a top performer in multiple public benchmarks, along with a proprietary model 
JAM~\cite{david2024} which has demonstrated strong results on real-world datasets and recent 
NIST evaluations~\cite{nist2024fatepdf}. Since the images in our study are synthetic and do not correspond to real individuals, we treat the \emph{prompted} age as the ground truth. Accordingly, in the experiments 
presented in the following sections particularly those involving MAE and other error metrics—the prompt age serves as the reference for evaluating the accuracy (or alignment) of the estimated ages.


\section{RESULTS AND DISCUSSION}
\label{sec:results}

\subsection{Comparison of Prompt Age vs. Estimated Age}

Table~\ref{tab:overall_metrics} summarizes the overall Mean Absolute Error (MAE), Root Mean Squared Error (RMSE), and Pearson correlation coefficient (Corr) across different T2I-model combinations. Lower MAE and RMSE indicate more accurate predicted ages. A Pearson correlation closer to 1 indicates the estimated ages track well with the changes in prompt age.

\begin{table}[ht]
\centering
\caption{Overall metrics for each \textbf{Age Estimator} and \textbf{T2I Model} combination. Lower MAE and RMSE, along with a Pearson correlation (Corr) closer to 1, all indicate stronger alignment between the prompt age and the estimated age.}
\label{tab:overall_metrics}
\begin{tabular}{llccc}
\hline
\textbf{Age Estimator} & \textbf{T2I Model} & \textbf{MAE} & \textbf{RMSE} & \textbf{Corr} \\
\hline
MiVOLO-D1 & Epic Realism & 2.99 & 3.96 & 0.984 \\
MiVOLO-D1 & Flux         & 5.26 & 6.73 & 0.959 \\
MiVOLO-D1 & SD 3.5       & 3.89 & 5.34 & 0.975 \\
JAM    & Epic Realism & 3.52 & 4.59 & 0.976 \\
JAM    & Flux         & 4.68 & 5.92 & 0.965 \\
JAM    & SD 3.5 & 3.58 & 4.92 & 0.986 \\
[0.5ex]
\hline
\end{tabular}
\end{table}

Overall, Epic Realism modes (for both MiVOLO-D1 and JAM) achieved relatively low MAE and high correlation, suggesting better alignment with the intended age. Conversely, \acs{Flux} exhibited higher MAE (5.26 for MiVOLO-D1, 4.68 for JAM), indicative of more pronounced age mismatch in the generated images.

\subsection{Analysis by Demographic Subgroups}
We further breakdown the performance by age groups, for example, 10-19, 20-29, etc., to examine how error varies between different age ranges. For brevity we highlight a few interesting examples. 

Teen and younger-adult prompts often show lower absolute errors, potentially because face-generating models systematically produce youthful-looking faces. Older age ranges (60+) appear more challenging, showing higher MAE and, in some cases, inconsistent correlation. In our age-bucket analysis, we focus on the JAM model due to its more consistent alignment with generated images overall. 

\paragraph{JAM + Flux}
For prompts aged 10--19, the correlation (0.83) is notably higher than 
in most other brackets, but the model still shows a sizable absolute error 
(MAE of 4.80). When examining younger-adult ranges like 20--29 and 30--39, 
the correlation drops sharply to 0.38 and 0.20, respectively, indicating that 
the model struggles to differentiate subtle changes within this demographic. 
Meanwhile, older age buckets (40--49 and 50--59) yield higher MAE (5.55--5.80) 
yet moderately improved correlation (0.65--0.51). In the 70--79 bracket, 
the MAE stands at 3.67 with a very low correlation (0.02), implying 
the model’s predictions remain misaligned at the extreme upper age ranges.

\paragraph{JAM + SD 3.5}
Here, teenage prompts (10--19) show the closest alignment, with an MAE of 1.92 
and a correlation of 0.64. As prompt ages increase to 30--39 or 40--49, 
the MAE climbs to approximately 4.2--4.4, though the correlation holds 
at a moderate 0.47--0.58, suggesting that while the model often overshoots 
or undershoots by a few years, it still tracks age variations to some degree. 
Performance deteriorates further in the 50+ brackets, where both the correlation 
(around 0.16--0.32) and error metrics degrade, indicating significant challenges in rendering older faces accurately.

\paragraph{JAM + Epic Realism}
Compared to \acs{Flux} and \acs{SD35}, \acs{EpicRealism} maintains relatively strong correlations 
(0.70--0.78) from the teenage years through the 40--49 bracket, though the corresponding 
MAE values can vary between 2.20 and 4.69. The biggest shortcoming emerges in the 
70--79 bracket, where the MAE reaches 7.69---the largest among all tested brackets---but 
the correlation remains at a modest 0.50. This outcome implies that while the system 
tends to deviate substantially from the correct numeric age in upper age ranges, it does so in a somewhat consistent manner, leaving room for improved calibration when generating more elderly faces.

When evaluating JAM on real-life production data, we observe a relatively consistent relationship between actual and predicted ages: although the model may occasionally overshoot or undershoot by a few years, it broadly tracks age variations across demographic brackets. This behavior manifests in moderate-to-high correlations combined with manageable MAE values. In contrast, when we examine synthetic data (across all three text-to-image generation modes), the results become substantially more erratic. Correlation and MAE sometimes reveal contradictory trends, indicating that, although the model can capture certain distinctions among age groups, it struggles to maintain uniformly accurate age estimates on synthetic faces. This discrepancy underscores the intricate challenges of creating truly photorealistic, age-appropriate synthetic images—and highlights that greater care and calibration are necessary before relying on such data for critical age estimation tasks.

\subsection{Finding Obvious Outliers}

We define an outlier as any image whose predicted age deviates by more than 
11 years from the prompt. When applying JAM to the \acs{Flux} or \acs{SD35} generation modes, we observe that roughly 2--3\% of the images appear 
``too young'' and another 3--5\% are ``too old'' (i.e., underestimating or 
overestimating the prompt age by over a decade). In contrast, MiVOLO-D1 consistently 
yields far more outliers: for instance, in the \acs{Flux} setting, more than 
8--10\% of generated faces fall outside the acceptable range on either side 
of the prompt age. These ``extreme mismatch'' samples typically present 
stylized or ambiguous facial features, suggesting that certain T2I pipelines 
struggle to preserve reliable age cues, particularly toward the lower and higher 
ends of the age spectrum.

\subsection{Bias \& Fairness Considerations}

To assess demographic disparities, we examined the MAEs by \textit{gender} and \textit{region}. With JAM + \acs{Flux}, males were on average estimated +2.18 years older 
than prompted, whereas females showed only +0.66. Conversely, JAM + \acs{SD35} skewed younger for both genders, but more so for females ($-1.10$ vs. $-0.26$ for males). Regional discrepancies were similarly apparent; for example, JAM + Epic Realism underestimated African prompts by about $-1.95$ and Asian prompts by $-2.62$ years, while often overestimating North American prompts.

Importantly, these biases appear \emph{more pronounced} in synthetic data than they do when the same estimator is applied to real-world faces. The T2I pipelines can amplify subtle disparities, especially if training data or facial embeddings lack balanced representation. Future investigations may thus consider additional calibration or more diverse synthetic datasets to mitigate systematic errors 
across demographic groups.

Overall, Epic Realism modes (both MiVOLO-D1 and JAM) typically exhibit lower average errors and higher correlations than \acs{Flux} or \acs{SD35}, but each generator has particular weaknesses—especially for older age prompts. The t-tests confirm that certain model–pipeline combinations differ significantly in performance, while others do not. Additionally, gender and region breakdowns uncover potential biases, as some groups’ ages are more overestimated or underestimated than others. Outlier inspection shows that extremely large age mismatches, while relatively rare, persist across all T2I settings—often for older faces or stylized prompts. Finally, regression analyses further corroborate these findings, revealing that, although correlation is often quite high, the error distribution can still be sizable for certain demographic slices.

\begin{figure}[ht!]
    \centering
    \includegraphics[width=0.22\textwidth]{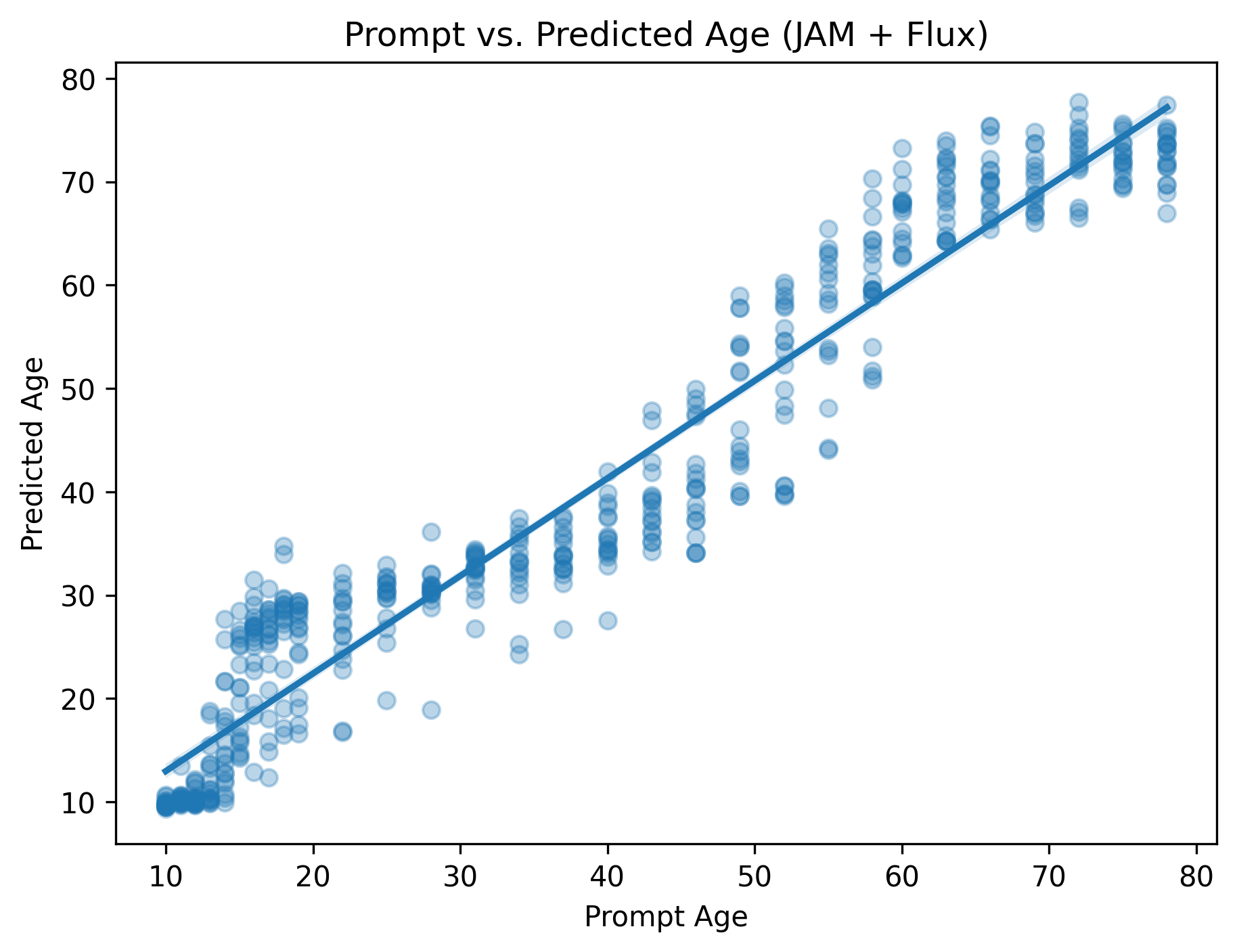}
    \hfill
    \includegraphics[width=0.22\textwidth]{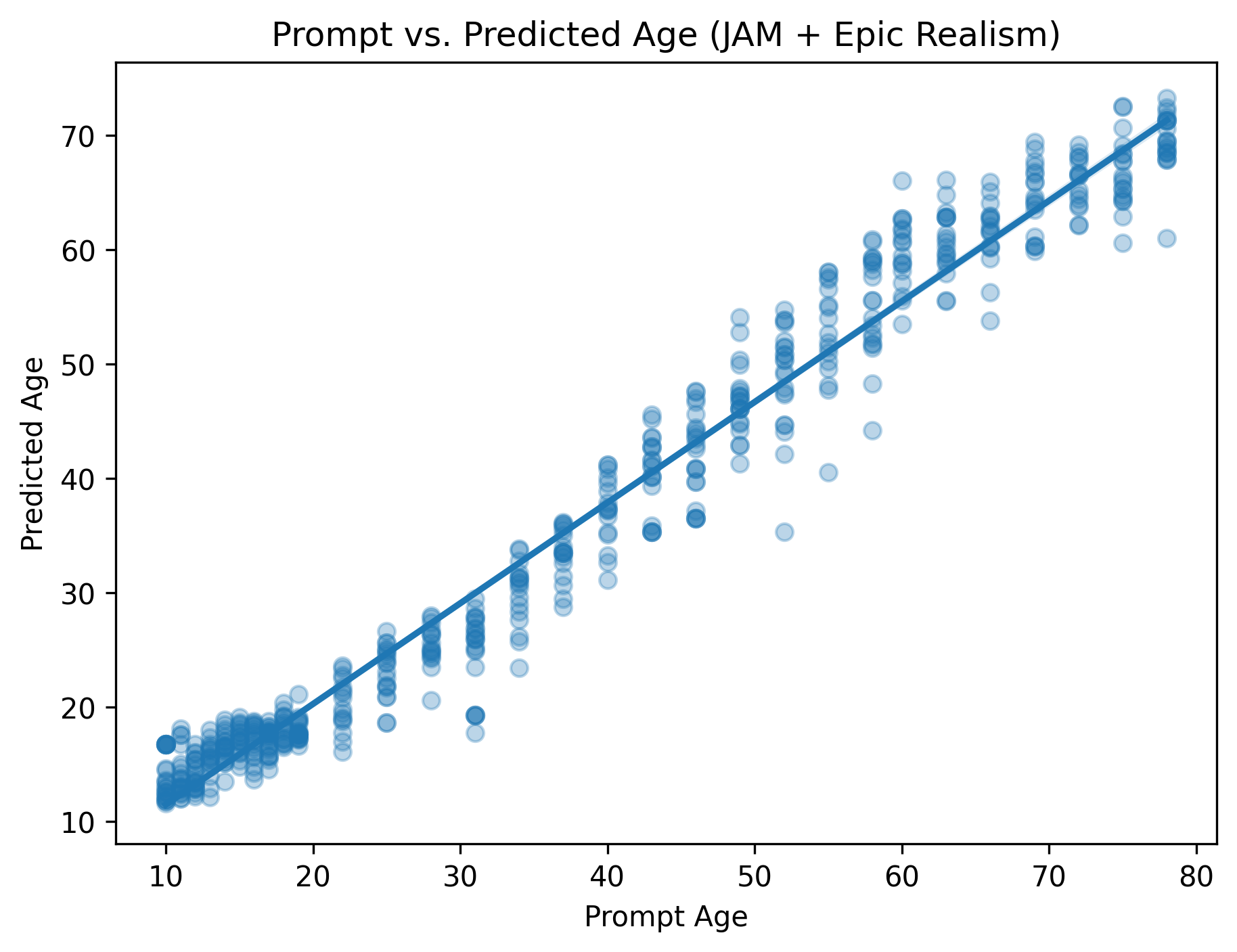}
    \caption{Prompt vs.\ Predicted Age plots with regression lines for 
    two T2I settings, Flux (left) and Epic Realism (right). Each point 
    represents a single synthetic face, and the diagonal line indicates 
    perfect age alignment.}
    \label{fig:regression_diagnostics}
\end{figure}

\subsection{Regression Diagnostics}
To further examine how well the predicted ages track the prompted ages, we plotted scatter diagrams of prompt age versus predicted age, overlaid with
a simple linear regression fit. Examples for two T2I settings, \acs{Flux} and \acs{EpicRealism}, are shown in Figure~\ref{fig:regression_diagnostics}.

Overall, \acs{EpicRealism} modes yield lines with slopes near 1.0, indicating high correlation, but exhibit a consistent offset that places many points either above or below the perfect-diagonal reference. In contrast, the \acs{Flux} setting displays a wider spread of data points, especially among younger and older age groups, in line with the higher MAE values observed. In summary, while correlation remains generally high across all tested settings, the distribution of residuals highlights systematic biases for
specific age brackets, emphasizing the need for careful filtering when precise age depiction is required.



\section{CONCLUSION}
\label{sec:conclusion}

This paper highlights the potential of modern text-to-image generative approaches 
for age-dependent tasks such as verification, data augmentation, and exploratory 
analysis in biometric pipelines. By systematically evaluating three state-of-the-art 
generative models against established age-estimation systems, we observed that while 
identity-related cues (e.g., face plausibility, general demographic traits) are 
often reasonably preserved, achieving \emph{precise} age depiction remains challenging. 
In many cases, our experiments revealed sizeable prediction errors or biases, 
particularly at the older or younger extremes of the prompt age range.


Overall, while generative AI shows promise for augmenting datasets in identity-centric contexts, its utility in producing age-accurate images remains limited. Practitioners must therefore weigh the risks of synthetic biases and mismatches, especially in high-stakes settings. Future work could focus on closing the gap between visually convincing outputs and semantically precise generations, potentially through more 
targeted training, improved prompt engineering, or post-hoc correction methods.


\section*{ETHICAL IMPACT STATEMENT}

\subsection{Risk}

We have not sought an external ethical review board for the age estimation aspects of this project, as both areas are well-established in the industry and are widely recognized as low-risk from an ethical standpoint. Age estimation is generally more accurately performed by modern ML/AI systems than by humans. AWS Rekognition demonstrates superior accuracy to humans in 9 out of 12 demographic categories~\cite{Ganel2022}. 

The project and the data have been reviewed internally by our company's ethics council and evaluated against our ethical policies. No violations of laws or ethical concerns have been identified.
One aspect to assess from a risk perspective is potential bias, as many age estimation systems exhibit bias issues. For example, age-related biases are a common concern in artificial intelligence systems.

One aspect to assess from a risk perspective is potential bias, as many age estimation systems exhibit bias issues. For example, age-related biases are a common concern in artificial intelligence systems~\cite{Chu2023}.

\subsection{Context of Age Estimation Modeling}

Although this work centers on evaluating and generating synthetic images, we employ age estimation modeling as part of our production environment. It is not the core focus of this study; however, we believe it is important to highlight how the age estimator is integrated and the care taken in its deployment. The insights and precautions described below reflect our commitment to ensuring fairness, reducing
bias, and safeguarding user privacy.

\subsection{Mitigation}

To mitigate bias, we employed curated datasets for model training and testing. These datasets are balanced across various age groups, genders, and countries of origin. However, as observed in NIST reports~\cite{nist2024fatepdf}, our solution still shows some biases—for instance, age estimation tends to be more accurate for males than for females (a trend observed across all participants and age groups), and middle-aged individuals (20–50 years) are more accurately assessed than older populations.

We hypothesize that age estimation accuracy may decrease in older populations because age-related changes in appearance become less pronounced, and gender bias could be influenced by cultural factors such as the use of cosmetics.

To address concerns regarding inappropriate use and to ensure transparency and accountability, our approach includes strict usage guidelines and the publication of detailed metrics that quantify the risks associated with model errors. These metrics are designed to provide clear insights into the system's performance across different demographics and scenarios, enabling continuous monitoring and improvement.

In our live traffic, we apply different mitigation strategies depending on the specific application:

\begin{itemize} 
    \item \textbf{Age Comparability}: Large age deltas are applied when making decisions to reject users in fraud detection applications. Such systems only act when a significant age mismatch is detected. 
    \item \textbf{Age Estimation}: Large confidence intervals are used, and age values are not directly applied or utilized in real applications due to the risk of errors. 
    \item \textbf{Age Verification}: Both large deltas and confidence intervals can be utilized. For instance, if a user attempts to access a restricted content website, the verification system rejects the users only when it is highly certain that the user is underage. 
\end{itemize}

These measures aim to minimize harm, provide transparency in model behavior, and maintain accountability across different age-sensitive use cases.

\subsection{Privacy and Security}

For all real images of individuals used in the research, we obtained direct consent from each user. When users utilize our applications for identity verification, we ask for their consent to use their data for commercial R\&D applications, including the development of age models. 

There is no direct compensation provided to the end user, but they benefit from improved service after certain development iterations. Only data from users above 18 years of age have been used, as evidenced by the government-issued ID documents associated with the data.

The image data is stored securely and encrypted, with eventual deletion in accordance with the retention policy or under data subject rights requests. No information or images that could lead to user identification are included in this manuscript. 

Additionally, the age data obtained from AI models is not stored and is removed immediately after analysis, ensuring that no sensitive information is retained.

\newpage
\bibliography{egbib}
\bibliographystyle{unsrt}

\end{document}